\pdfoutput=1

\ifdefined \IEEESTYLE
    \documentclass[lettersize,journal]{IEEEtran}
\else
    \documentclass{article}
    \usepackage{arxiv}
\fi

\newif\ifshowcomments
\showcommentsfalse

\usepackage{amsmath,amsfonts}
\usepackage{siunitx} %
\usepackage{algorithmic}
\usepackage{algorithm}
\usepackage{array}
\usepackage{subcaption} %
\usepackage{textcomp}
\usepackage{stfloats}
\usepackage{url}
\usepackage{verbatim}
\usepackage{graphicx}
\usepackage{lipsum}
\usepackage[table,svgnames]{xcolor}
\usepackage{tcolorbox}
\usepackage{fontawesome5} %
\usepackage{transparent}
\usepackage[acronym]{glossaries} %
\usepackage{hyperref}
\usepackage{fancyvrb}
\usepackage{placeins}
\hypersetup{hidelinks} %

\usepackage[backend=biber,style=ieee]{biblatex}

\newcommand{\passat}[1]{\mbox{\texttt{pass@#1}}}

\date{}

\newacronym{ai}{AI}{Artificial Intelligence}
\newacronym{arc}{ARC}{Abstraction and Reasoning Corpus}
\newacronym{dsl}{DSL}{Domain-Specific Language}
\newacronym{ea}{EA}{Evolutionary Algorithm}
\newacronym{ec}{EC}{Evolutionary Computation}
\newacronym{ge}{GE}{Grammatical Evolution}
\newacronym{llm}{LLM}{Large Language Model}
\newacronym{lrm}{LRM}{Large Reasoning Model}
\newacronym{nn}{NN}{Neural Network}
\newacronym{tta}{TTA}{Test-Time Adaptation}
\newacronym{ttt}{TTT}{Test-Time Training}

\begin{document}

\ifdefined \IEEESTYLE
    \title{Toward a Symbolic Solver for the Abstraction and Reasoning Corpus Using Evolutionary Computation}
    \renewcommand{\shorttitle}{Fairly Symbolic}
\else
    \title{\texorpdfstring{\smash{\makebox[0pt][l]{\transparent{0}\resizebox{262pt}{!}{\normalfont Ceci n'est pas une pipe}}}\includegraphics[width=262pt]{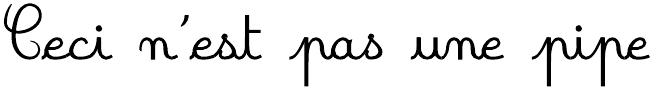}}{Ceci n'est pas une pipe} \\[-10pt] Toward a Symbolic Solver for the Abstraction and Reasoning Corpus Using Evolutionary Computation}
    \renewcommand{\shorttitle}{Ceci n'est pas une pipe (Fairly Symbolic)}
\fi

\ifdefined \IEEESTYLE
    \author{
      Luca Favole, Giorgio Mongardi, Giovanni Squillero,~\IEEEmembership{Senior Member,~IEEE}, and Alberto Tonda
    
      \thanks{L. Favole, G. Mongardi, and G. Squillero are with Politecnico di Torino, Torino, Italy.}
      \thanks{A. Tonda is with Universit\'{e} Paris-Saclay, INRAE, Applied Mathematics and Computer Science Lab (UMR 518 MIA-PS), 91120, Palaiseau, France; and with CNRS, Complex Systems Institute of Paris \^{I}le-de-France (UAR 3611 ISC-PIF), 75013, Paris, France.}
      \thanks{All authors contributed equally to this work and their names are listed in alphabetical order.}
    }
    
    \markboth{IEEE Transactions on Evolutionary Computation,~Vol.~XX, No.~Y, Month~Year}%
    {Shell \MakeLowercase{\textit{et al.}}: A Sample Article Using IEEEtran.cls for IEEE Journals}
    
    \IEEEpubid{0000--0000/00\$00.00~\copyright~2021 IEEE}
\else
    \author{ \href{https://orcid.org/0009-0009-5688-7388}{\includegraphics[scale=0.06]{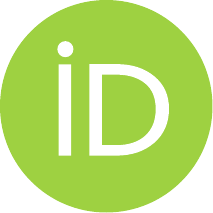}\hspace{1mm}Luca~Favole} \\
	Department of Control and Computer Engineering\\
	Politecnico di Torino\\
	Corso Duca degli Abruzzi 24, 10129, Torino, Italy\\
	\texttt{luca.favole@polito.it} \\
	\And
	\href{https://orcid.org/0009-0009-6175-2947}{\includegraphics[scale=0.06]{orcid.pdf}\hspace{1mm}Giorgio~Mongardi} \\
    Department of Control and Computer Engineering\\
	Politecnico di Torino\\
	Corso Duca degli Abruzzi 24, 10129, Torino, Italy\\
	\texttt{giorgio.mongardi@polito.it} \\  
	\And
	\href{https://orcid.org/0000-0001-5784-6435}{\includegraphics[scale=0.06]{orcid.pdf}\hspace{1mm}Giovanni~Squillero} \\
    Department of Control and Computer Engineering\\
	Politecnico di Torino\\
	Corso Duca degli Abruzzi 24, 10129, Torino, Italy\\
	\texttt{giovanni.squillero@polito.it} \\
	\And
	\href{https://orcid.org/0000-0001-5895-4809}{\includegraphics[scale=0.06]{orcid.pdf}\hspace{1mm}Alberto~Tonda} \\
    Applied Mathematics and Computer Science\\
	Université Paris-Saclay, INRAE\\
	22 place de l'Agronomie, 91120, Palaiseau, France\\
    Complex Systems Institute of Paris\\
    CNRS\\
    113 rue Nationale, 75013, Paris, France\\
	\texttt{alberto.tonda@inrae.fr} \\
    }

\hypersetup{
pdftitle={Toward the Abstraction and Reasoning Corpus with Symbolic Evolutionary Computation},
pdfsubject={cs.LG}, %
pdfauthor={Luca~Favole, Giorgio~Mongardi, Giovanni~Squillero, Alberto~Tonda},
pdfkeywords={Abstraction and reasoning corpus, ARC-AGI, artificial intelligence, evolutionary algorithms, genetic programming, symbolic reasoning},
}
\fi

\maketitle

\ifdefined \IEEESTYLE
\else
    \makeatletter
    \begingroup
    \renewcommand{\thefootnote}{}
    \long\def\@makefntext#1{\noindent #1}
    \footnotetext{All authors contributed equally to this work and their names are listed in alphabetical order.}
    \footnotetext{\copyright~2026 The Authors. This work is licensed under a Creative Commons Attribution 4.0 International License (\href{https://creativecommons.org/licenses/by/4.0/}{CC~BY~4.0}).}
    \endgroup
    \makeatother
\fi

\begin{abstract}
The Abstraction and Reasoning Corpus (ARC) is a benchmark designed to assess the capability of Artificial Intelligence (AI) systems to generalize starting from a few samples, a task humans excel at. The ARC-AGI contests held since 2020 have been the focus of intense competition, with participants proposing approaches ranging from traditional symbolic AI to Transformer architectures. Despite a long history of successes, evolutionary algorithms (EAs) were notably absent from the approaches appearing in the main contest. As score-and-search program synthesis approaches, EAs suffer from issues related to choosing the proper representation in the Domain-Specific Language applied to the problem (for example, pixel-level or object-level), and distinguishing between two candidate solutions producing the same outputs. In this work, we present the first fully symbolic, evolutionary-based program synthesis approach submitted to the ARC-AGI-2 main contest. The main contributions are: evolving the grid parser and the output projector together with the program that transforms the grid, so that a task-specific language emerges during program synthesis; grounding the symbols the parser can extract in core knowledge theory from cognitive psychology; and an object-aware fitness that grades partially correct outputs beyond pixel-level discrepancies.
\end{abstract}

\ifdefined \IEEESTYLE
\fi

\section{Introduction}
\label{sec:introduction}
\ifdefined \IEEESTYLE
    \IEEEPARstart{W}{ith} %
\else
    With %
\fi
the recent advances in \gls{ai} mainly driven by \glspl{llm} and \glspl{lrm}, it has become more and more necessary to design benchmarks able to assess the real capabilities of algorithms. The \gls{arc} was originally proposed in 2019~\cite{chollet2019measure} and later used in hackathons (2020--2023) and competitions (2024--2026)~\cite{chollet2025arcprize,chollet2026arcprize}. It quickly gathered considerable attention: hundreds of teams of \gls{ai} researchers and practitioners attempted to outperform each other, and large companies volunteered to test their latest \glspl{llm} and \glspl{lrm} on the benchmark. The techniques adopted by participants range from symbolic program synthesis~\cite{ferre2024tackling} to purely neural~\cite{franzen2025product}, to neuro-symbolic approaches~\cite{bednarek2024learning}. Despite a few early attempts at employing \gls{ec}~\cite{fischer2020solving}, and some more recent approaches labeled as ``evolutionary'', yet basically wrapping \glspl{llm}~\cite{pourcel2025selfimproving,berman2024substack}, \glspl{ea} have been notably absent from \gls{arc} competitions.

In this work, we present FairlySymbolic, an evolutionary approach to solve the problems in \gls{arc}. It is, to the best of our knowledge, the only \gls{ea} which competed on the ARC-AGI-2 benchmark and was scored on the \emph{semi-private evaluation set} of the 2026 edition of the ARC Prize. Building upon the preliminary ideas presented in~\cite{favole2026gp}, FairlySymbolic evolves concurrently the code for a \emph{parser}, to extract symbols from the images; a \emph{program}, to manipulate such symbols and compute the solution; a \emph{projector}, to reconstruct the output image from the solution. 

The symbols are extracted from images by interpreting bitmaps using a small number of \textit{a priori} concepts, inspired by the \emph{core knowledge} theory by Spelke and colleagues~\cite{spelke2000coreknowledge,spelke2007coreknowledge,spelke2024whatbabiesknow}. Taking inspiration from~\cite{xu2023graphs}, symbols are internally encoded as the nodes and edges of a graph. This format simplifies the synthesis of programs, since all operations act on a uniform type of object. Different tasks in \gls{arc}, however, need different types of symbols: some consider individual pixels and their color, while others identify objects on a background. The symbols available to the program constitute the \gls{dsl} of that specific task. As Wittgenstein wrote in 1921, ``the limits of my language mean the limits of my world''\footnote{\emph{Die Grenzen meiner Sprache bedeuten die Grenzen meiner Welt.}}~\cite[5.6]{wittgenstein1921logisch}: a rich \gls{dsl} is essential to find the solution, yet, the search space of program synthesis grows rapidly with the number of available symbols, so, to keep the problem tractable, the \gls{dsl} should be as small as possible. In the proposed approach, the parser is optimized, that is, the \gls{dsl} is defined, while performing the program synthesis: only the relevant symbols will be extracted. 

As stated above, in FairlySymbolic a candidate solution is a triplet composed of a parser, a program, and a projector. Checking whether a solution transforms the input of a demonstration pair into the correct output is trivial. The \gls{ea}, however, also needs to compare imperfect solutions and decide which one is closer to the target, so the fitness function must grade partial success. The evaluation depends on the symbols in the task \gls{dsl}. For example, if the problem requires symbols for \emph{shapes} and \emph{directions}, a bitwise comparison is likely to be useless. The proposed evaluation considers properties such as object counts, matching quality, colors, shapes, and spatial geometry, providing more informative feedback for partially correct solutions and effectively guiding the evolutionary search, an approach which is also inspired by core knowledge.

The main contributions of our work are the following:

\begin{itemize}
    \item We present an evolutionary algorithm able to tackle \gls{arc} problems by concurrently optimizing the code for a parser, a program, and a projector --- that is, evolving a task-specific \gls{dsl} during program synthesis.
    \item Drawing on cognitive psychology and linguistics, we propose a well-founded definition of the innate symbols needed to tackle problems like those in \gls{arc}.
    \item We analyze a fitness evaluation strategy that goes beyond exact matching and pixel-level discrepancies by incorporating object-level information. 
\end{itemize}

The rest of the paper is organized as follows: Section~\ref{sec:arc} describes the \gls{arc} benchmarks and summarizes the state-of-the-art solutions; Section~\ref{sec:core-knowledge} introduces the core knowledge theory and the innate priors it provides; Section~\ref{sec:approach} details the proposed approach, from the parsing of grids into typed multigraphs to the fitness function and the evolutionary search; Section~\ref{sec:experiments} reports the experimental evaluation on ARC-AGI and ConceptARC; finally, Section~\ref{sec:discussion} discusses the results and outlines future work.
\FloatBarrier

\section{The Abstraction and Reasoning Corpus}
\label{sec:arc}

\gls{arc} is a series of benchmarks originally proposed by François Chollet~\cite{chollet2019measure} to evaluate the effectiveness of \gls{ai} systems on problems requiring generalization starting from a few samples --- a task which humans can typically do with ease. Aiming at measuring the ability to adapt to ``unknown unknowns across a broad category of related tasks'', \gls{arc} consists of few-shot abstraction and analogy problems, and represents a discontinuity in the evaluation of \gls{ai} systems. The benchmark immediately exposed important limitations of \glspl{nn} as its design prevented well-known pitfalls, such as data contamination and approximate retrieval. At the time of writing, more than five years after the first competition, human performance on this benchmark remains steadily well above that of AI systems.

Initially \gls{arc} distinguished between a \emph{public training set}, consisting of \num{400} tasks openly accessible for researchers and developers to build, analyze, and train their solvers; a \emph{public evaluation set}, consisting of another \num{400} tasks openly accessible for local validation and checking generalization on unseen tasks; and \num{200} non-public tasks, held out by the organizers to evaluate solvers fairly and prevent data contamination. In 2024, this set was further split in \num{100} \emph{semi-private tasks} for evaluating models that require external API calls or unconstrained internet/test-time compute access, and \num{100} {private tasks}, reserved strictly for official leaderboard scoring on isolated, offline competition infrastructure within strict compute limits. 

ARC-AGI-2 updated the structure with a \emph{public training set} of \num{1000} tasks for training; a \emph{public evaluation set} with \num{120} tasks for assessing the performances; a \emph{semi-private evaluation set} with \num{120} tasks for benchmarking frontier or proprietary AI systems that require internet connectivity, external API access, or extended test-time compute budgets; and a \emph{private evaluation set} with \num{120} tasks, strictly confidential, and reserved exclusively for official competition leaderboard scoring on isolated, offline execution infrastructure under strict time and compute limits (12 hours of offline CPU computation or 12 hours on four L4 GPUs, and \qty{1}{\giga\byte} of hard disk, to solve 120 hidden tasks).

\begin{figure*}[h]
    \centering
    \begin{subfigure}{0.99\textwidth}
        \centering
        \includegraphics[width=0.8\linewidth]{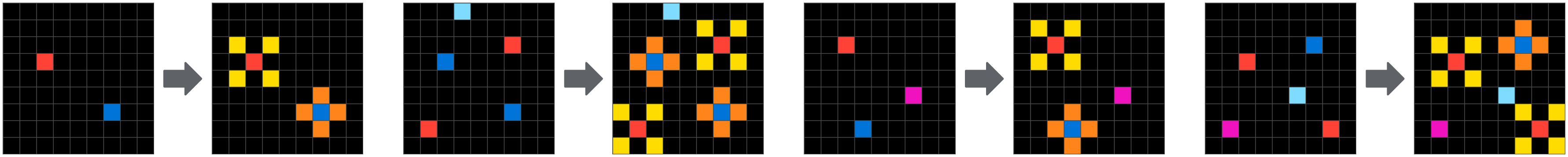}
        \caption{ARC-AGI-1, easy training task \texttt{0ca9ddb6} ($9\times9$ grids). One fixed rule: each red cell gains yellow diagonal neighbors, and each blue cell gains orange orthogonal neighbors.}
        \label{fig:arc-agi-examples-1}
    \end{subfigure}\\[2ex]
    \begin{subfigure}{0.99\textwidth}
        \centering
        \includegraphics[width=\linewidth]{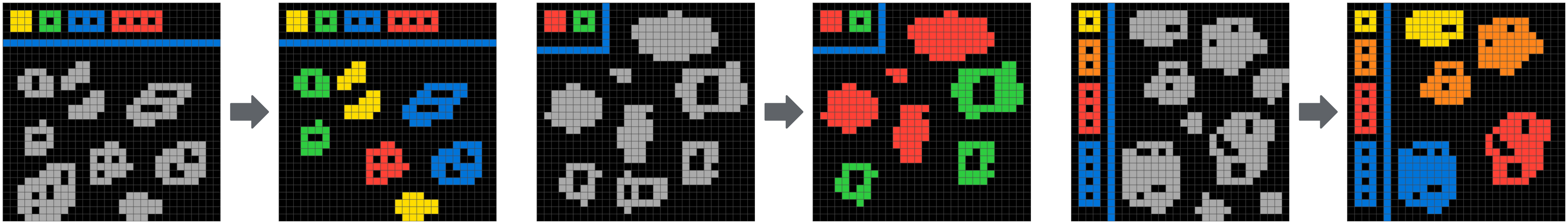}
        \caption{ARC-AGI-2, harder public evaluation task \texttt{e3721c99}, second test pair ($30\times30$ grids). In-context symbols: the legend blocks pair colors with hole counts, and each gray shape takes the matching color or vanishes.}
        \label{fig:arc-agi-examples-2}
    \end{subfigure}
    \caption{Example tasks: Each arrow leads from an input grid to its output grid; in each row, the first pairs are the demonstration pairs and the last is the test pair. Sources: \url{github.com/fchollet/ARC-AGI} and \url{github.com/arcprize/ARC-AGI-2}.\label{fig:arc-agi-examples}}
\end{figure*}

\subsection{State-of-the-art Solutions}
\label{ssec:state-of-the-art}

Surveying the literature on \gls{arc} can be quite challenging, as participants make their code public, but often do not summarize their approaches in reports. Furthermore, comparing approaches is also not straightforward, as some techniques have only been tested on ARC-AGI-1 and others just on ARC-AGI-2. Still, it is possible to gather a good amount of information from participants who described their work, either in scientific literature or in blog posts, and from the ARC Prize technical reports of 2024 and 2025~\cite{chollet2025arcprize,chollet2026arcprize}. 

Over the course of the different editions of the \gls{arc} competition, it is possible to identify general trends of proposed approaches. Early years (2020--2022) were dominated by purely symbolic approaches, often exploiting program synthesis based on clever exhaustive enumeration constrained by program length~\cite{ferre2024tackling}. 

The best-known evolutionary baseline for \gls{arc} is an attempt to use \gls{ge} over a \gls{dsl} by Fischer et al.~\cite{fischer2020solving}. It defines a \gls{dsl} of image operations and uses \gls{ge} to generate \gls{dsl} expressions, mapping codons to syntax trees. Evolved expressions that reproduce the demonstration pairs serve as the task solver. On the test tasks, a random-search baseline could not solve a single task, so evolutionary pressure did help, though absolute performance was modest.

\Glspl{nn} first entered program search as guides: an adaptation of the DreamCoder framework~\cite{ellis2023dreamcoder} to the ARC-AGI-1 benchmark~\cite{alford2022neuralguided} used a learned model to steer the enumeration of programs. The 2023 competition then saw the first successful Transformer-based entries, which fine-tuned the \gls{nn} weights on each task at inference time, a technique later popularized as \gls{ttt} or \gls{tta}~\cite{chollet2025arcprize,franzen2025product}. When the competition was relaunched in 2024 by the newly created ARC Prize Foundation, the state-of-the-art private score rose from \qty{33}{\percent} to \qty{55.5}{\percent}, and the vast majority of the participants employed neural architectures. Besides \gls{ttt}, other neural approaches used a combination of program synthesis and direct generation of the output grids~\cite{li2025combining}.

After the end of the 2024 competition, the impressive results obtained by the novel \glspl{lrm} like OpenAI's o3 led to the opening of the secondary ranking computed on the separate semi-private test set, where users were able to employ any kind of technique, far less restricted by time or computational effort. In this second bracket, participants could experiment with proprietary \gls{llm} and \gls{lrm} APIs from commercial providers, such as OpenAI and Anthropic. Here we observe remarkable attempts at employing techniques that their creators termed ``evolutionary'':

Berman created Evolutionary Test-Time Compute, \emph{Python} and \emph{natural-language} versions. Both versions exploit \glspl{llm} as mutation and crossover operators, but the former evolves Python functions, while the latter, English instructions. The approach is effective: the first version scored \qty{53.6}{\percent} on ARC-AGI-Pub with Claude 3.5 Sonnet, the public record at the time~\cite{berman2024record}; the second scored \qty{29.4}{\percent} on ARC-AGI-2, and \qty{79.6}{\percent} on ARC-AGI-1 at about \$8.42 per task~\cite{berman2025english,dong2025evolutionary} using Grok-4. Berman's work served as a starting point for later \gls{llm}-evolutionary systems.
    
Pourcel, Colas, and Oudeyer's SOAR (self-improving evolutionary program synthesis) solves \qty{52}{\percent} of the ARC-AGI-1 public test set~\cite{pourcel2025selfimproving}. It alternates between an evolutionary search, where an \gls{llm} samples and refines candidate programs, and a hindsight-learning phase that turns those attempts into problem--solution pairs used to fine-tune the \gls{llm}'s sampling and refinement abilities. It won second place in the ARC Prize 2025 paper awards~\cite{knoop2025results}.

Pang combines Berman-style \gls{llm} evolution with DreamCoder's idea of growing a reusable library, scoring \qty{77.1}{\percent} on ARC-AGI-1 and \qty{26.0}{\percent} on ARC-AGI-2, both on the semi-private evaluation set~\cite{pang2025efficient,pang2025code}. Promising programs are added to an expanding library, and the current best program from the library is included in the \gls{llm}'s prompt. This addresses a weakness of earlier systems, which treated each task as an isolated problem.

Imbue's \emph{Code Evolution} runs a population-based evolution of Python solutions for each task~\cite{mewes2026beating}. Each ``organism'' is a piece of Python code: a parent is sampled in proportion to its fitness, an \gls{llm} mutates it into children, and the scored children are added back to the population. Imbue contrasts this with consensus-based selection and with Berman's approach, using global population-level sampling with a more sophisticated per-organism scoring mechanism and emphasizing divergence over convergence.

Dyachenko's \emph{LLM-seeded genetic search} uses the \gls{llm} only once, then lets a genetic algorithm do the rest~\cite{dyachenko2026combining}. The \gls{llm} writes programs in a small \gls{dsl}, these seed the initial population, and the genetic algorithm evolves them toward a solution with no further \gls{llm} calls. A fixed-width \gls{dsl} keeps mutated programs valid and lets the system execute \num{8000} candidate programs per generation on a GPU. On 60 public ARC-AGI-2 evaluation tasks, the \gls{llm} solved 2 outright and genetic search added 4 more, for 6 in total.

\FloatBarrier

\subsection{Core Knowledge}
\label{sec:core-knowledge}

The theory of \emph{core knowledge}, developed in cognitive science, proposes that human cognition builds upon a set of fundamental systems that emerge early in development and provide the foundations for more complex reasoning and learning. Spelke and Kinzler~\cite{spelke2007coreknowledge} describe four main systems concerned with objects, actions, numbers, and space, while also considering a fifth system related to social cognition. Subsequent work has expanded this framework to six systems~\cite{spelke2024whatbabiesknow}.

In designing ARC, Chollet drew upon this theory, identifying a set of elementary cognitive priors intended to guide the design of its tasks. These include \emph{objectness}, which enables the recognition of distinct entities and their properties; \emph{goal-directedness}, which supports reasoning about agents, actions, and intended outcomes; \emph{numbers and counting}, which provide elementary quantitative reasoning; and \emph{basic geometry and topology}, which encompass spatial relationships, connectivity, containment, and geometric transformations. By grounding the tasks in these fundamental concepts, ARC aims to provide a fairer basis for comparing human and artificial intelligence, reducing dependence on acquired knowledge and previous experience~\cite{chollet2019measure}.

For artificial intelligence, this perspective highlights two complementary challenges. The first is to develop fundamental structures that allow an agent to interpret its environment and extract meaningful information from it. These structures should also provide a foundation upon which more complex knowledge can be built through learning. The second challenge concerns learning itself: developing mechanisms capable of building upon these foundations to acquire, combine, store, and apply new knowledge. In symbolic approaches, abstractions and \glspl{dsl} attempt to provide such foundations by representing entities, properties, relationships, and the operations that can be performed on them.
\FloatBarrier

\subsection{ConceptARC}
\label{ssec:conceptarc}

The importance of basic concepts in abstract reasoning also motivates evaluating how well \gls{ai} systems can understand and use specific concepts to solve problems. For this purpose, Moskvichev et al.~\cite{moskvichev2023conceptarc} introduced ConceptARC, retaining the grid-based format of ARC while organizing tasks into 16 concept groups targeting specific spatial and semantic concepts, such as counting, object extraction, spatial relationships, and shape completion. Each group contains 10 tasks with varying levels of complexity and abstraction. Each task provides one to four training examples and three test inputs, for a total of 160 tasks and 480 test inputs. Unlike ARC-AGI, where solving a task may require combining multiple concepts, ConceptARC focuses each task on one or a few specific concepts, enabling a more targeted assessment of a solver's conceptual capabilities and helping identify systematic weaknesses that may be difficult to distinguish from aggregate performance on ARC-AGI.
\FloatBarrier

\section{Proposed Approach}
\label{sec:approach}

\subsection{Joint Evolution of Representations and Programs}
\label{ssec:approach-overview}

For each ARC task, the solver searches for an executable solution that reproduces the provided input--output pairs and can subsequently be applied to the test input. The configurations that the evolutionary search can explore for the parser and projector define the \emph{parser configuration space} and \emph{projector configuration space}, respectively, while the \emph{domain-specific language} (DSL) defines the symbolic operations available for constructing programs. The parser converts an input grid into a typed multigraph; the program transforms that graph; and the projector reconstructs the predicted output grid from the transformed graph. The three components are evaluated together against all the available input--output pairs.

The joint search is motivated by the observation that the transformations required to solve a task depend on how the input is represented. For instance, moving a shape is straightforward when the shape is represented as a single object, whereas the same transformation may require several operations if that shape is decomposed into individual cells or multiple objects. Conversely, grouping disconnected regions of the same color into a single object can obscure their individual identities when they need to be manipulated independently. An effective representation therefore exposes the relevant entities and relationships while avoiding unnecessary complexity.

Rather than fixing the abstraction beforehand, the proposed approach evolves the parser together with the symbolic program operating on the extracted graph. This allows the search to discover representations in which the required transformations become expressible within the available language, or in which otherwise complex transformations can be achieved through simpler sequences of symbolic operations. The projector is also included in the joint search, since output reconstruction can affect the correctness of a solution, particularly when objects overlap, the background changes, or the output dimensions differ from those of the input.

\subsection{Parsing the Grid into a Typed Multigraph}
\label{ssec:approach-parser}

The parser constructs a directed typed multigraph from each input grid, where nodes represent entities identified in the scene, carrying type-specific attributes, and edges encode relationships between them. The basic entities are represented by \texttt{SubObject} nodes, corresponding to objects extracted from the grid and characterized by their occupied cells, color, mass, bounding box, centroid, and shape descriptors. Additional node types represent higher-level or implicit structures identified through further analysis, such as clusters and cavities. As a multigraph, the representation allows multiple typed edges to connect the same pair of nodes, capturing different relationships simultaneously, such as two objects being adjacent while also sharing a row or having the same shape. The resulting representation preserves the geometric information required to reconstruct the grid, while providing the transformation program with direct access to the entities, their attributes, and their relationships.

\subsubsection{Object Extraction}

Graph construction begins with background identification, using either an explicitly specified color or a heuristic based on color distribution, corners, and grid boundaries. Depending on the configuration, background cells can be excluded from object extraction or treated as part of the scene. The parser can also identify structural grid lines that partition the scene into zones. These are detected by examining horizontal and vertical lines for sufficiently consistent non-background colors, according to a configurable threshold. Recognized separators are represented as structural objects, while the enclosed regions become \texttt{ZoneNode}s. Importantly, separator pixels are excluded from ordinary object extraction, preventing them from connecting otherwise separate regions.

The remaining cells are grouped into objects according to configurable color and connectivity criteria. Adjacent cells can form monochromatic or multicolored objects, with four-neighbor connectivity considering only horizontal and vertical adjacency, and eight-neighbor connectivity also including diagonal contact. The extracted objects can subsequently be refined through configurable operations, including decomposition into simpler geometric components, merging based on containment-related criteria, and splitting according to the spatial relationships between objects. Additional filtering criteria can also be applied to exclude or reclassify selected components.

\begin{figure*}[!t]
    \centering
    \includegraphics[width=\textwidth]{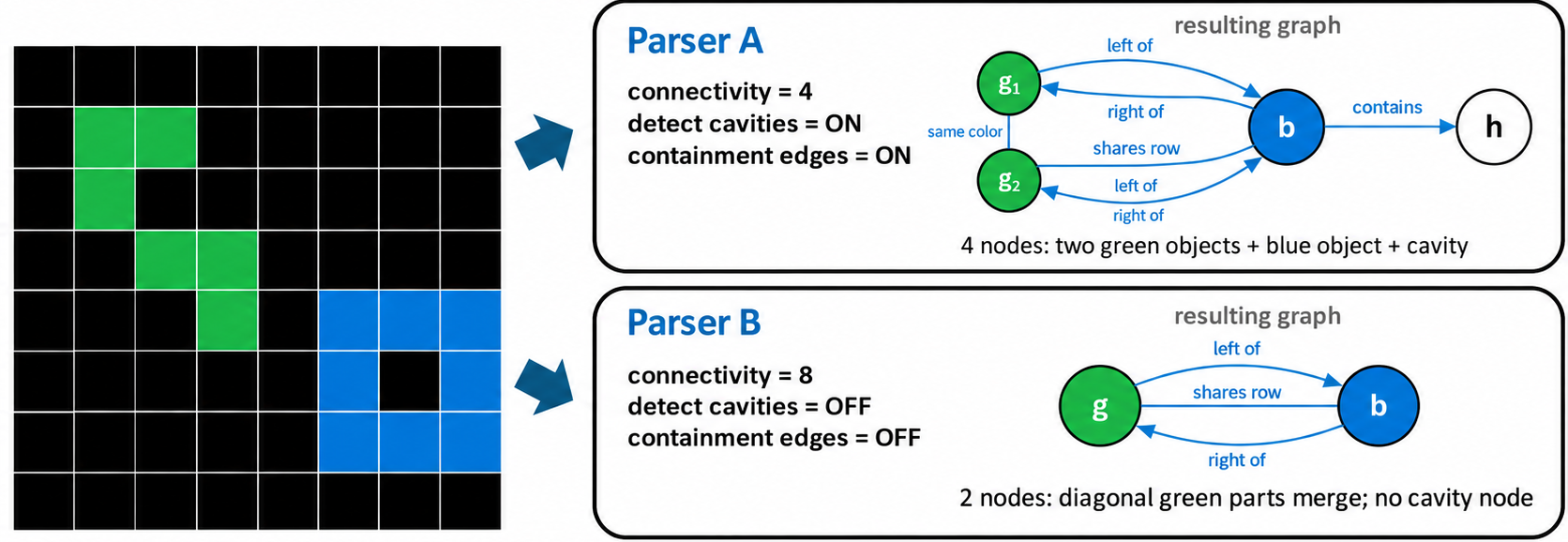}
    \caption{Different graph representations of the same grid. Parser A uses four-neighbor connectivity and explicitly detects the cavity enclosed by the blue object. Parser B uses eight-neighbor connectivity and omits cavity detection. Consequently, the two green objects merge into one and the cavity node disappears. Only the nodes and relations relevant to this example are shown.}
    \label{fig:parser-contrast}
\end{figure*}

Fig.~\ref{fig:parser-contrast} provides a simple illustrative example of how different parser configurations can produce distinct graph representations of the same input grid.

\subsubsection{Additional Structures and Relations}

Following object extraction, the parser can perform additional analysis passes to identify higher-level structures and enrich the graph with contextual information. Rather than evaluating every possible structure and relationship, the parser executes the analyses enabled by its configuration, which also specifies the criteria and thresholds used by each detector. Cavities are identified as enclosed regions of background cells through a flood-fill procedure, while clusters are formed by grouping adjacent objects into larger, potentially multicolored structures. Repeated patterns are detected by identifying objects with matching colors and shapes arranged at regular spatial intervals, whereas symmetries are recognized by testing the grid against horizontal, vertical, and diagonal reflections.

Detected structures are incorporated into the graph through additional nodes, such as \texttt{CavityObject}, \texttt{ClusterObject}, and \texttt{PatternObject}, while implicit structures, such as symmetry axes, can be represented without corresponding occupied cells. These entities coexist with the original objects, preserving their individual properties while exposing higher-level arrangements. Additional analyses compute geometric and contextual attributes, including local coordinates, zone membership and features, spatial ordering, and noise-related characteristics.

Relationships are established by evaluating geometric and logical conditions between entities, generating typed edges whenever the corresponding conditions are satisfied. Spatial relations describe contact, overlap, and relative position, while containment relations connect objects with enclosing structures or their constituent parts. Other relation families capture alignment, similarities in color or shape, symmetries, ray projections, gravity-like directions, and bridges between objects. Multiple relationships can coexist between the same entities, allowing symbolic programs to access objects according to their structural roles and relationships rather than reconstructing these properties through pixel-level operations.

\subsubsection{Symbolic Programs}

Transformations are expressed through a \gls{dsl} operating on the graph representation. A program consists of an ordered sequence of rules, optionally followed by a global transformation. Each rule combines a selector, an action, and optional guards: the selector identifies the entities to manipulate, the guards determine whether the action should be executed, and the action specifies the transformation to perform. Unlike rule-level actions, global transformations operate on the scene as a whole, supporting operations such as rotations, reflections, cropping, and tiling without requiring individual objects to be selected and manipulated separately. The \gls{dsl} provides multiple selector types, action types, and three types of count-based guards.

Selectors primarily identify \texttt{SubObject} nodes according to their attributes, structural properties, or relationships with other entities in the graph. This allows selection criteria to exploit information derived from higher-level structures without necessarily manipulating those structures directly. Attribute-based selectors consider properties such as color, shape, mass, position, and boundary contact, while comparative selectors identify entities according to relative properties, such as the largest, smallest, leftmost, or rightmost object. Relational selectors exploit the graph structure to retrieve objects sharing the shape of a reference, touching objects of a given color, or connected through specific edge types. More complex selections can be constructed through nested queries and set operations, including union, intersection, and difference. For instance, these mechanisms allow a program to select the smallest object among those satisfying a particular structural relationship.

Guards impose additional conditions on the selected set, constraining the minimum, maximum, or exact number of entities returned by the selector. When guards are specified, the action is executed only if all conditions are satisfied; otherwise, it is applied directly to the selected entities.

Actions modify the selected entities through operations such as recoloring, deletion, retention, translation, reflection, copying, scaling, and geometric manipulation. More elaborate operations support filling cavities or bounding boxes, drawing outlines, completing symmetries, extending lines, constructing regular patterns, and performing transformations based on object counts, spatial arrangements, or interactions with other objects. Each action can take parameters defining the transformation, such as displacement values for translation or color and connectivity settings for outlining. By combining selection, conditional execution, and manipulation, the language expresses interpretable transformations that explicitly identify the entities involved and the operations performed on them.

\subsubsection{Sequential Execution}

Rules are executed sequentially, with each rule operating on the scene produced by the preceding operations. Their order is therefore significant, as earlier transformations can modify the objects available to subsequent rules. For example, a first rule may select the objects belonging to the least frequent color and remove all other objects from the scene. A second rule can then select the surviving objects and draw an outline around them. The result of the second operation therefore depends on the selection and transformation performed by the first.

Since transformations can modify the structure of the scene, some attributes and relationships computed during parsing may become outdated. To maintain consistency, the executor can reconstruct the graph between consecutive rules through a rendering and re-encoding cycle. When an action has been executed and the next selector requires information that may no longer be valid, the intermediate scene is converted into a grid using a fixed rendering procedure. This initializes the canvas with the scene background (or zero if unavailable) and paints the objects in their existing order, overwriting previously occupied cells in case of overlap, without applying postprocessing. The resulting grid is then parsed again using the individual's parser configuration, reconstructing the objects, their attributes, and their relationships from the modified scene. Selectors relying only on attributes maintained directly by the executed actions can instead operate on the existing graph, avoiding unnecessary reconstruction.

When present, a global transformation is applied after the sequence of symbolic rules. Depending on the operation, it can act directly on the transformed scene or on its rendered grid. When the program contains no rules, certain global transformations can instead be applied directly to the original input grid, bypassing the graph-based execution. This allows global transformations to express whole-grid operations independently or complement the modifications performed by preceding rules.

\subsubsection{Output Reconstruction}

Unlike the fixed renderer used for intermediate graph reconstruction between rules, the projector is a configurable component of the individual responsible for converting the transformed scene into a grid used to produce the final prediction. It operates after the symbolic rules and, depending on the global transformation, before or after its application. The projector configuration space covers four policies: background initialization, rendering order, overlap handling, and optional postprocessing. It initially creates a canvas matching the dimensions of the current scene, which may have been modified by preceding transformations. More generally, the output dimensions can differ from those of the input, as transformations applied to the scene or directly to the rendered grid can expand or reduce them. The canvas is initialized using the detected background, zero, or the most frequent color in the input or transformed scene. Alternatively, the projector can start from a copy of the input grid when its dimensions are compatible.

Objects are rendered in their original order or sorted according to properties such as mass, position, or depth. When objects overlap, the selected policy determines whether later objects overwrite previously rendered cells or preserve existing non-background values. Optional postprocessing can restore structural grid lines or crop the output to the bounding box of its foreground, removing unnecessary margins. Consequently, different projector configurations can produce different predictions even when the parser and symbolic program remain unchanged, making projection an integral part of the evolutionary search.

\subsection{Evolutionary Search}
\label{ssec:approach-search}

\subsubsection{Fitness Evaluation and Candidate Ranking}
\label{sec:approach-fitness}

The fitness evaluates how accurately a candidate reproduces the demonstration outputs. A demonstration pair is considered solved only when the predicted grid matches the target exactly, including its dimensions and every cell value. A candidate fits the task when all demonstration pairs are reproduced exactly. While exact matching determines whether a solution has been found, a graded distance measure provides feedback for imperfect predictions, guiding the evolutionary search toward increasingly accurate transformations.

The main distance measure combines object-level comparison with pixel-level disagreement. To evaluate objects consistently across candidates, the predicted and target grids are independently decomposed into monochromatic four-connected components, excluding the most frequent color of each grid as its estimated background. This extraction procedure is independent of the individual's evolved parser, allowing the fitness to compare predictions using a common representation regardless of the abstractions employed by the candidate program.

Objects from the predicted and target grids are greedily paired in descending order of pixel intersection-over-union (IoU), with unmatched objects contributing to the matching error. Five complementary components are then computed: the discrepancy in object counts~$d_n$, the quality of object matching~$d_m$, the color disagreement between matched objects~$d_c$, the difference in their translation-invariant shapes~$d_s$, and the difference in bounding-box position and dimensions~$d_b$. These components are normalized and combined into the object distance~$d_o$

\begin{equation}
d_o =
\min\left(1,\,
\frac{d_n + 2d_m + d_c + d_s + 0.5d_b}{5.5}
\right).
\label{eq:arc-object-distance}
\end{equation}

The pixel-level distance~$d_p$ measures the number of mismatched cells, including discrepancies caused by different grid dimensions. It is normalized by the number of non-zero cells in the target, with an alternative normalization for entirely zero-valued targets. For an imperfect prediction, the combined distance is

\begin{equation}
d = d_o + 0.3\,d_p.
\label{eq:arc-soft-distance}
\end{equation}

Both distances, $d_o$ and~$d_p$, are bounded between zero and one. The object-level component provides feedback on partially correct object configurations, while the pixel-level component retains sensitivity to individual cell errors. Exact predictions are assigned zero distance, and a purely pixel-based evaluation mode is also supported.

The distances computed across the demonstration pairs are used to rank candidates lexicographically. Candidates reproducing all demonstration outputs exactly take precedence over imperfect solutions. Among imperfect candidates, the ranking first favors a larger proportion of exactly reproduced pairs, followed by a smaller maximum distance across the demonstration pairs and, subsequently, a smaller mean distance. Program complexity serves as the final tie-breaking criterion.

For candidates reproducing every demonstration pair, program complexity instead becomes the primary ranking criterion. It accounts for the number of rules and their arguments, favoring simpler programs among equally correct solutions. This ranking determines which candidates are preserved through elitism, identifies the best individual, and provides the comparison criterion for subsequent refinement procedures.

\subsubsection{Evolutionary Process}

The search begins with an enumeration of simple candidate solutions, rather than relying exclusively on randomly generated programs. This preliminary stage explores individual global transformations, short symbolic programs, and selected combinations of rules, together with alternative parser and projector configurations. For the parser, six representative configurations provide different initial choices for object extraction, background handling, and structural and relational analyses. The generated candidates are evaluated against the demonstration pairs, and promising solutions are retained to initialize the evolutionary population alongside randomly generated individuals. This provides the search with potentially useful transformations while preserving diversity for tasks requiring more elaborate programs.

At each generation, the population is evaluated on the demonstration pairs and ordered according to the ranking criteria described above. The highest-ranked individuals are preserved through elitism, while the remaining population is generated through parent selection and variation operators. Parent selection is performed separately from the overall ranking, using $\varepsilon$-lexicase selection. Rather than comparing candidates through a single aggregate score, this mechanism considers multiple error cases independently. For each demonstration pair, these include the five object-distance components, pixel mismatch, color-specific errors, and errors measured over four spatial regions of the output.

During $\varepsilon$-lexicase selection, the error cases are considered in randomized order. At each step, candidates are retained if their error on the current case does not exceed the best observed error by more than a tolerance computed using the median absolute deviation of that case's errors across the population. Filtering continues until only one candidate remains or all cases have been considered, in which case a survivor is selected randomly. This allows individuals performing particularly well on specific aspects of a task to contribute to subsequent generations, even when they are not among the highest-ranked candidates overall. Tournament selection is also supported as an alternative.

Variation operates on complete individuals, modifying their parser configurations, symbolic programs, and projectors. The parser is governed by a set of parameters controlling the different stages of graph construction. These parameters are evolved alongside the symbolic program and projector configuration. Parser mutations modify one or two of these parameters, potentially affecting object extraction, background handling, and structural detection. Program mutations can replace selectors or actions, modify their parameters and guards, insert, remove, or reorder rules, and change the global transformation. Projector mutations modify background initialization, overlap handling, or postprocessing policies.

Crossover combines the configurations and programs of two parents. The evolvable parser parameters are inherited independently from either parent, while the remaining configuration is inherited from one parent. The symbolic program is recombined by selecting cut points in the parents' rule sequences and joining the resulting segments, subject to the maximum program length. The global transformation and projector configuration are inherited from either parent. Since the parser has no predefined target graph, changes to its configuration are evaluated indirectly through the predictions produced by the complete individual, allowing the search to explore different combinations of representations and transformations.

To maintain diversity, the search avoids introducing duplicate symbolic programs within the same generation and introduces additional mutations or randomly generated individuals when progress stagnates. Multiple independent restarts further explore different regions of the search space. Fitness caching and the reuse of previously parsed input scenes reduce redundant computations, particularly when candidates share configurations or are evaluated repeatedly. The best solutions discovered during the evolutionary process are subsequently considered for local refinement.

\subsubsection{Post-Evolution Refinement}

After each evolutionary run, the best candidate may undergo a sequence of local refinement procedures, subject to the remaining computational budget. Unlike the population-based evolutionary search, these procedures explore restricted modifications of promising solutions, accepting changes according to the previously defined ranking criteria. The first stage applies a greedy hill-climbing procedure based on mutations of the current individual. Each mutated candidate is evaluated against the demonstration pairs and retained only if it improves the ranking. The procedure terminates when a complete solution is found, a maximum number of attempts is reached, or progress stagnates.

Numerical refinement subsequently examines the highest-ranked individuals from the final evolutionary population. For each candidate, the procedure systematically evaluates alternative values for supported numerical arguments of selectors and actions, including colors, object counts, thresholds, and spatial offsets. Each modification is evaluated independently, and improvements are retained before examining subsequent arguments. This allows the search to adjust parameters of promising programs without changing their overall rule structure.

An additional error-guided refinement procedure is applied when a candidate is sufficiently close to reproducing the demonstration outputs. It analyzes the differences between predicted and target grids to identify recurring color discrepancies. These observations guide mutations toward potentially useful corrections, such as modifying the output color of an existing action or introducing a recoloring rule. The resulting candidates are evaluated through a local hill-climbing procedure, with the error information updated whenever an improvement is accepted.

Program compression then attempts to remove rules individually, retaining each deletion when the resulting candidate is not worse under the ranking criteria. The process is repeated until no further rule can be removed, reducing unnecessary instructions while preserving or improving performance on the demonstration pairs. Unlike the preceding refinement procedures, compression is also applied to candidates that already reproduce every demonstration pair. If the best candidate remains imperfect, the system additionally evaluates alternative projector configurations, systematically varying background initialization, overlap handling, and postprocessing while keeping the rendering order fixed.

\subsubsection{Post-Solution Search and Final Selection}
\label{sec:post-solution-search}

When an evolutionary restart produces a candidate that reproduces all demonstration pairs exactly, the solver skips the remaining scheduled restarts and redirects the unused computational budget to post-solution search. It first attempts to find simpler alternatives through additional evolutionary searches, starting with a maximum number of rules one lower than that of the shortest exact-fit program found so far. Whenever a shorter solution is found, the limit is reduced again and another search is performed. This process stops when no solution is found under the new limit, the limit cannot be reduced further, or the remaining budget is exhausted. The candidates discovered through these additional searches are retained alongside those obtained from the completed restarts and their refinement procedures.

Once these searches are completed, the solver gathers and deduplicates the candidates obtained from the initial restarts, refinement procedures, and searches for shorter programs, and performs a final ranking to select the primary solution. For exact-fit candidates, this ranking incorporates an additional generalization heuristic. For each demonstration pair, the heuristic identifies input colors absent from all other demonstration inputs and temporarily removes rules whose selector or guard arguments contain integer values matching those colors. The modified program is then reevaluated on all demonstration pairs, and a penalty is assigned if it no longer solves every pair. This penalty takes precedence over program complexity, with an optional selector-specificity criterion providing an additional preference for less specialized programs. The highest-ranked candidate is selected as the primary solution, whose prediction initially constitutes the first attempt.

With the primary solution selected, the solver can search for an alternative prediction by exploring transformed views of the task, including transposition and horizontal or vertical reflection. Each transformation is applied to both the input and output grids of the demonstration pairs, and a separate evolutionary search is performed in the transformed space. Successful programs are then evaluated on the correspondingly transformed test input, and their predictions are mapped back to the original orientation. The most frequently produced prediction differing from the primary one is selected as the second attempt. If at least two view-based solutions agree on this alternative and none supports the primary prediction, the attempts are swapped, promoting the consensus prediction to the first position. If the transformed views produce no distinct alternative, the second attempt is instead obtained from a diversified pool of candidate solutions.
\FloatBarrier

\section{Experiments and Results}
\label{sec:experiments}

\subsection{Experimental Setup}
\label{ssec:program}

FairlySymbolic is written in C++17, with no external dependencies and no neural component. It is distributed on GitHub under an \emph{MIT No Attribution} (MIT-0) permissive license\footnote{\url{https://github.com/EvoAGI/FairlySymbolic}}. Except for the official private-set submissions, all experiments were conducted on a machine equipped with a 36-thread Intel Xeon Gold 6140 processor. Each task is processed by a single-threaded search, with up to 17 tasks running concurrently. Computational budgets are expressed as wall-clock time per task rather than as numbers of candidate evaluations. Unless otherwise specified, the evolutionary search uses a population of 40 individuals, five restarts sharing the total per-task time budget, and programs of at most five rules.

The parser configuration space, \gls{dsl}, and projector configuration space were initially designed manually, based on general considerations about the representations and transformations potentially useful for ARC tasks, without targeting specific problems or concept groups. We subsequently used ConceptARC to identify limitations in the capabilities available to the solver and introduced additional functionality to address them. In the current implementation, the \gls{dsl} provides 55 action types, with evolvable arguments where applicable; the parser is governed by 44 parameters; and the projector has four configuration parameters. These extensions were introduced pragmatically to cover the concepts represented in ConceptARC, rather than through a systematic optimization of the available operations and configurations. We used the resulting near-complete coverage of ConceptARC as a development milestone before returning to the evaluation on ARC-AGI.

ARC-AGI comprises public \emph{training} and \emph{evaluation} sets, as well as hidden \emph{semi-private} and \emph{private} evaluation sets. Each task provides \emph{demonstrations} (input--output pairs) and one or more \emph{test inputs} whose outputs must be predicted. Ground-truth test outputs are publicly available for the training and public evaluation sets, allowing offline assessment, whereas the semi-private and private sets are evaluated through official submissions, with the former used for the public leaderboard and the latter for the final competition ranking. In all cases, the search has access only to the demonstrations and test inputs, never to the ground-truth test outputs. A task is considered \emph{fitted} when a candidate solution reproduces all demonstration outputs, and \emph{solved} at \passat{k} when at least one of its first $k$ predictions is correct for every test input. The official ARC Prize protocol allows up to two predictions per test input. For ConceptARC, \passat{k} is instead reported over individual test inputs.

\subsection{ARC-AGI}
\label{ssec:results-arc-agi}

Table~\ref{tab:run-coverage} shows that our system finds correct solutions for 219 of the 400 ARC-AGI-1 training tasks across the available runs. Table~\ref{tab:main} provides a more detailed breakdown of the results on the public ARC-AGI-1 and ARC-AGI-2 splits. On ARC-AGI-1, our system solves 110 of 400 public evaluation tasks at \passat{2} with \qty{3840}{\s} per task. The difference between the training and evaluation sets suggests that their task distributions are not equally well covered, but it is distinct from the question of generalization within a task. There, the important distinction is between a candidate that reproduces the demonstrations and one that also predicts the withheld test outputs: in the public evaluation set, 164 tasks are fitted, but only 110 are solved at \passat{2}. These counts do not directly measure overfitting, since fitted and solved tasks do not necessarily coincide. Furthermore, increasing the search budget yields considerably more fitted tasks than additional correct predictions. Overall, these results demonstrate that a purely symbolic evolutionary approach can solve a substantial fraction of ARC-AGI-1 without neural training or pretrained representations, and suggest that further optimization of the \gls{dsl} and configuration spaces could potentially saturate the benchmark.

\begin{table}[!t]
\caption{Comparison between the best single run and cumulative task coverage across all available runs at \passat{2}.\label{tab:run-coverage}}
\centering
\begin{tabular}{lrr}
\hline
Benchmark & Best run & All runs \\
\hline
ARC-AGI-1 training   & 197/400 & 219/400 \\
ARC-AGI-2 evaluation &   3/120 &   3/120 \\
\hline
\end{tabular}
\end{table}

\begin{table}[!t]
\caption{Fitted tasks (Fit.) and tasks solved at \passat{1} and \passat{2} with 840
and \qty{3840}{\s} per task: the best of three seeds at \qty{840}{\s}, one seed at
\qty{3840}{\s}. Of the \num{1000} ARC-AGI-2 training tasks, 767
come from ARC-AGI-1; the last row covers the 233 that do not.\label{tab:main}}
\centering
\setlength{\tabcolsep}{4pt}
\begin{tabular}{lrrrrrrr}
\hline
 & & \multicolumn{3}{c}{\qty{840}{\s}} & \multicolumn{3}{c}{\qty{3840}{\s}} \\
Benchmark & Tasks & Fit. & @1 & @2 & Fit. & @1 & @2 \\
\hline
ARC-AGI-1 training   &  400 & 237 & 182 & 193 & 258 & 183 & 197 \\
ARC-AGI-1 evaluation &  400 & 146 &  96 & 103 & 164 & 103 & 110 \\
ARC-AGI-2 training   & 1000 & 405 & 273 & 294 & 455 & 283 & 304 \\
ARC-AGI-2 evaluation &  120 &   7 &   3 &   3 &  10 &   3 &   3 \\
\hline
ARC-AGI-2 tr., new   &  233 &  42 &  13 &  20 &  55 &  19 &  20 \\
\hline
\end{tabular}
\end{table}

On ARC-AGI-2, the limitation becomes more pronounced. Its training set includes 767 tasks inherited from ARC-AGI-1, so the 304/\num{1000} training result at \passat{2} is not representative of performance on entirely new ARC-AGI-2 problems. On the 233 new training tasks, our system solves 20 (\qty{8.6}{\percent}), whereas on the public evaluation set it solves 3 of 120 (\qty{2.5}{\percent}). This is consistent with the greater demand for combinations of operations and more elaborate abstractions in ARC-AGI-2~\cite{chollet2025arcagi2}. In particular, increasing the nominal budget yields additional fits on ARC-AGI-2 evaluation without improving the number solved, which cautions against treating search time alone as the main bottleneck.

\ifdefined \IEEESTYLE
  \providecommand{\resultsfigwidth}{\columnwidth}
\else
  \providecommand{\resultsfigwidth}{0.6\textwidth}
\fi

\begin{figure}[!h]
\centering
\includegraphics[width=\resultsfigwidth]{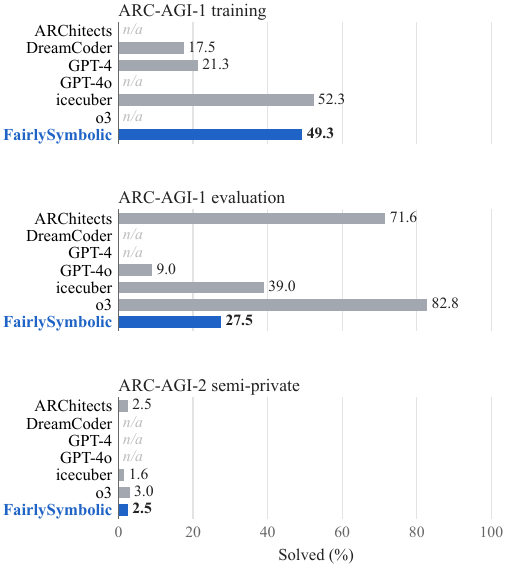}
\caption{Comparison of the percentage of tasks solved on ARC-AGI with published systems. Our system, FairlySymbolic, was limited to \qty{3840}{\s} per task on ARC-AGI-1; on ARC-AGI-2 the figure reports the result of the Kaggle entry. All systems were allowed two attempts per task, except on ARC-AGI-1 training, where icecuber, GPT-4, and DreamCoder used three attempts; and on ARC-AGI-1 evaluation, where GPT-4o used one attempt. It should also be noted that, on ARC-AGI-1 evaluation, the results for o3 refer to a \emph{pre-release} version using six samples per task, and those for the ARChitects to the \emph{2025 method}.}
\label{fig:compare-arcagi}
\end{figure}

\begin{figure}[!t]
\centering
\includegraphics[width=\resultsfigwidth]{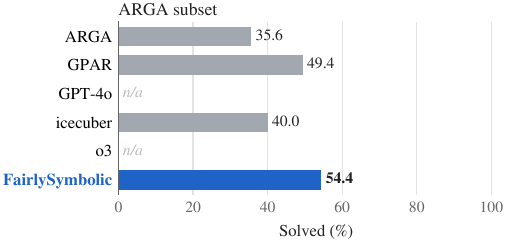}
\caption{Percentage of tasks solved at pass@1 on the 160-task ARC-AGI-1 subset used by ARGA. Our system used 1\,800\,s per task.}
\label{fig:compare-arga}
\end{figure}

Fig.~\ref{fig:compare-arcagi} compares our results with those of other approaches to ARC-AGI. On ARC-AGI-1, our system achieves results comparable to some established symbolic approaches, particularly icecuber, which also relies on a manually designed \gls{dsl} and symbolic search. ARGA~\cite{xu2023graphs} and GPAR~\cite{lei2024generalized} provide further relevant comparisons, as both employ object-based representations and program synthesis. On the 160-task ARC-AGI-1 subset used in ARGA's experiments, our system solves 87 tasks at \passat{1}, compared with 57 reported by ARGA and 79 verified from GPAR's published solutions, positioning our approach among the leading symbolic program synthesis methods for ARC-AGI.

The gap becomes substantial when considering the strongest approaches, such as the ARChitects and o3, which achieve considerably higher scores on ARC-AGI-1. Nevertheless, our approach obtains its results without neural networks or pretrained models, relying entirely on evolutionary search and manually designed symbolic capabilities. This distinction is relevant when interpreting its performance. The limitations become particularly apparent on ARC-AGI-2, where the system solves only \qty{2.5}{\percent} of the semi-private evaluation tasks, suggesting that the representations, compositionality and transformations currently available are insufficient for many of the more demanding problems.

\begin{table}[!t]
\caption{Fitted ARC-AGI-1 training tasks that are also solved at \passat{2}, by the
number of rules in the program (\qty{120}{\s} per task, summed over three seeds).\label{tab:rules}}
\centering
\begin{tabular}{lrrrrr}
\hline
Rules & 0 & 1 & 2 & 3 & $\ge$4 \\
\hline
Fitted & 90 & 296 & 103 & 80 & 62 \\
Solved & \qty{100}{\percent} & \qty{95}{\percent} & \qty{77}{\percent} & \qty{64}{\percent} & \qty{47}{\percent} \\
\hline
\end{tabular}
\end{table}

The number of symbolic rules provides another view of generalization. Table~\ref{tab:rules} shows that almost all programs with zero or one rule that fit the demonstrations also solve the corresponding tests, whereas this fraction decreases markedly for longer programs, reaching \qty{47}{\percent} with four or more rules. This does not establish that additional rules cause overfitting: more difficult tasks may require longer explanations, and a larger program space admits more accidental fits. It does show that the solver becomes less reliable at distinguishing competing explanations as the expressions it can construct become more elaborate. This pattern is particularly relevant when considering the still limited results on ARC-AGI-2, where stronger compositional capabilities are likely to be needed without simply increasing the number of unconstrained rule combinations.

\begin{figure}[!t]
\centering
\includegraphics[width=\resultsfigwidth]{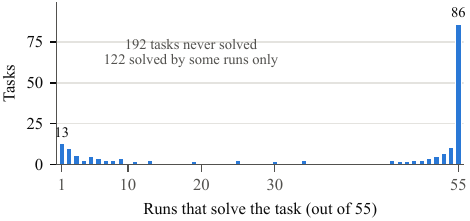}
\caption{For each of the 400 ARC-AGI-1 training tasks, the number of runs that
solve it at \passat{2}, over the 55 runs of the parameter study (earlier version of
the solver, \qty{120}{\s} per task) that allow programs of at least three rules.}
\label{fig:frequency}
\end{figure}

Fig.~\ref{fig:frequency} separates tasks that are consistently solvable from those sensitive to the particular search run and parameter configuration. Of the 400 ARC-AGI-1 training tasks in the 55-run parameter study, 86 are solved in every run, 122 in only some, and 192 in none. The large always-solved and never-solved groups suggest that the performance of the explored configurations is not determined by random variation, while the intermediate group shows that some opportunities remain in the search process. In addition, the never-solved tasks cannot be classified as \gls{dsl} limitations without checking whether their intended solutions are expressible.

\begin{table}[!t]
\caption{Weaker searches on the ARC-AGI-1 training tasks at \qty{120}{\s} per task:
fitted and solved (\passat{2}) tasks and median generations per task, best of
three seeds (the text compares seed means).\label{tab:fitness}}
\centering
\begin{tabular}{lrrr}
\hline
Search & Fitted & Solved & Generations \\
\hline
Full fitness          & 212 & 184 & \num{2600} \\
Coarse fitness        & 158 & 153 & \num{3900} \\
Random search         &  98 & 107 & \num{3300} \\
Random, no refinement &  74 &  92 & \num{3500} \\
\hline
\end{tabular}
\end{table}

The ablation in Table~\ref{tab:fitness} tests whether the graded fitness and evolutionary search contribute beyond drawing many random programs. Replacing the full object- and pixel-aware fitness with a coarse score that counts only exactly reproduced demonstrations reduces \passat{2}, despite the greater number of generations made possible by cheaper evaluations. Randomly generating candidates instead of applying evolutionary variation reduces successful predictions further; removing the local refinement stage produces another decline. These results indicate that the quality of feedback, guided variation, and refinement each contribute to the search. The fact that the random-search baselines can solve more test tasks than they fit on the demonstrations also reinforces why the difference between fitted and solved totals must not automatically be equated with the number of overfitted solutions.

\begin{figure}[!t]
\centering
\includegraphics[width=\resultsfigwidth]{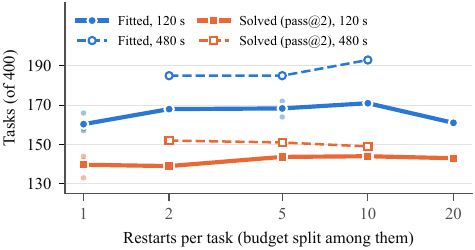}
\caption{Fitted and solved ARC-AGI-1 training tasks against the number of
restarts (earlier version of the solver). Solid: \qty{120}{\s} per task, three seeds (dots) for 1 and 5 restarts, one
otherwise. Dashed: \qty{480}{\s} per task, one seed.}
\label{fig:restarts}
\end{figure}

The restart study in Fig.~\ref{fig:restarts} examines how distributing a fixed computational budget across multiple searches affects performance. The results indicate that a moderate number of restarts provides a good balance between search diversity and the time available to each search, with the default choice of five restarts achieving consistently competitive results. At the same time, Table~\ref{tab:run-coverage} shows that independent runs can discover solutions to different tasks, increasing the overall coverage from 197 to 219 tasks on ARC-AGI-1 training. Together, these observations highlight the benefits of search diversification while suggesting that performance is relatively robust to the exact number of restarts.

\subsection{ConceptARC}
\label{ssec:results-conceptarc}

As discussed in the experimental setup, we used ConceptARC as a development benchmark, extending the available symbolic capabilities until they covered almost all of it, before returning to ARC-AGI. Consequently, the almost perfect results reported in Table~\ref{tab:conceptarc-concepts}, reaching \qty{99.4}{\percent} at \passat{2}, are not surprising, as the extensions were specifically designed to address the limitations identified on this benchmark. Returning to ARC-AGI after these improvements, we observed substantial gains: on ARC-AGI-1, solved tasks increased from 164 to 197 on training and from 73 to 110 on public evaluation, while on ARC-AGI-2 training they increased from 247 to 304. Performance on ARC-AGI-2 public evaluation, however, remained unchanged at 3 solved tasks. Overall, this experience highlights the usefulness of ConceptARC in identifying missing conceptual capabilities in symbolic systems built around core knowledge, providing concrete guidance for extending their representations and operations.

\begin{table}[!t]
\caption{Percentage of ConceptARC test inputs solved at \passat{2}, before and after the extensions.\label{tab:conceptarc-concepts}}
\centering
\setlength{\tabcolsep}{6pt}
\begin{tabular}{lrr}
\hline
Concept & Before & After \\
\hline
TopBottom2D    & 70.0 & 100.0 \\
SameDifferent  & 66.7 & 96.7 \\
ExtractObjects & 56.7 & 100.0 \\
\multicolumn{3}{c}{$\vdots$} \\
CompleteShape  & 13.3 & 96.7 \\
Order          & 6.7 & 100.0 \\
Count          & 6.7 & 100.0 \\
\hline
\textbf{Average} & \textbf{36.9} & \textbf{99.4} \\
\hline
\end{tabular}
\end{table}

\begin{figure}[h]
\centering
\includegraphics[width=\resultsfigwidth]{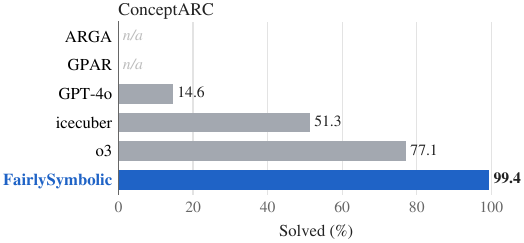}
\caption{Percentage of ConceptARC test inputs solved. FairlySymbolic used two attempts with 16 voted searches (600\,s per task in total), icecuber used three attempts, and the remaining systems used one.}
\label{fig:compare-conceptarc}
\end{figure}

\subsection{Qualitative Analysis}

\begin{figure*}[!t]
\centering
\begin{minipage}[t]{0.32\textwidth}
    \centering
    \includegraphics[width=\linewidth]{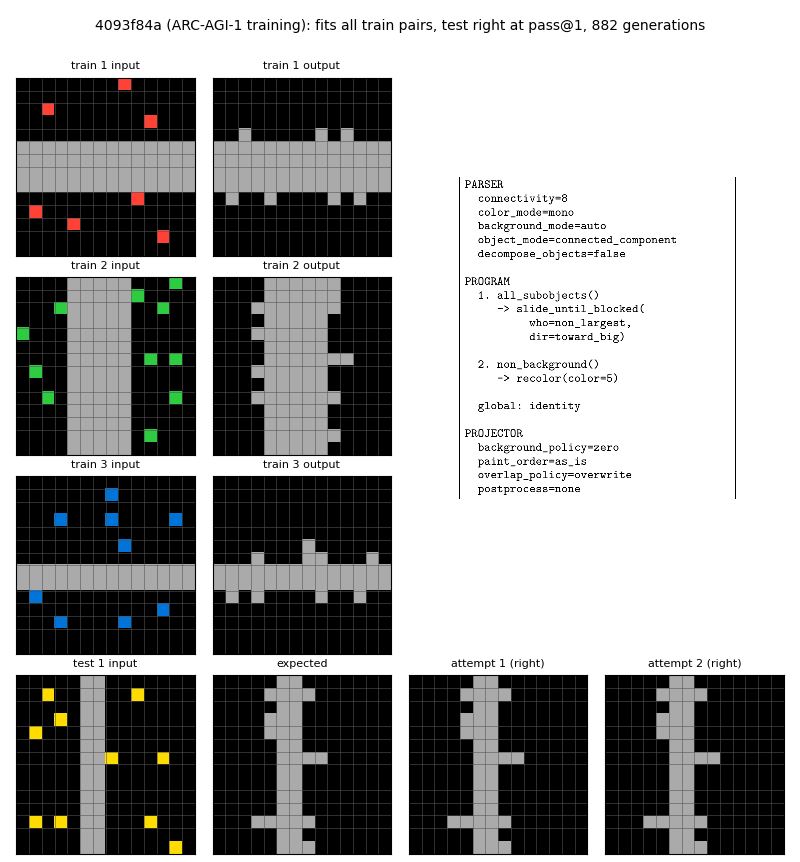}
    \par\smallskip
    \textbf{(a)} Successful solution
\end{minipage}
\hfill
\begin{minipage}[t]{0.32\textwidth}
    \centering
    \includegraphics[width=\linewidth]{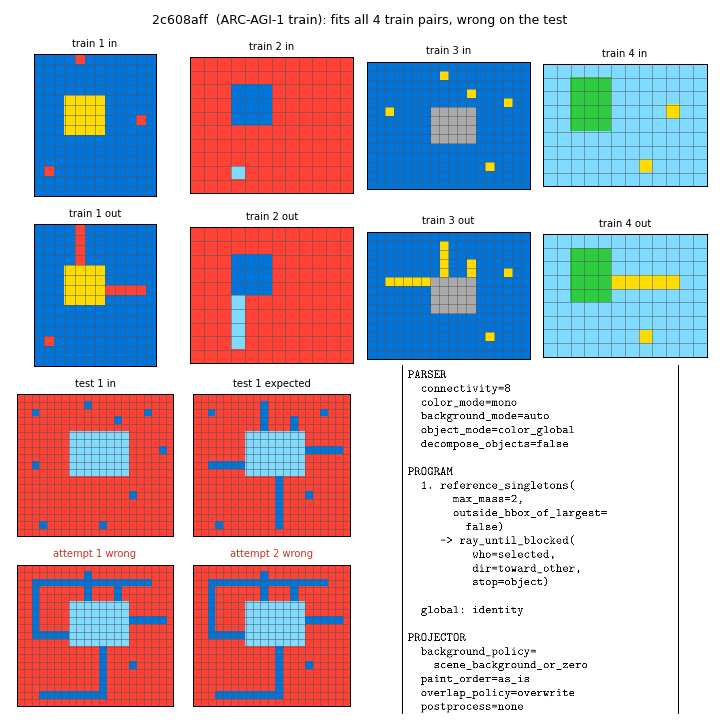}
    \par\smallskip
    \textbf{(b)} Generalization failure
\end{minipage}
\hfill
\begin{minipage}[t]{0.32\textwidth}
    \centering
    \includegraphics[width=\linewidth]{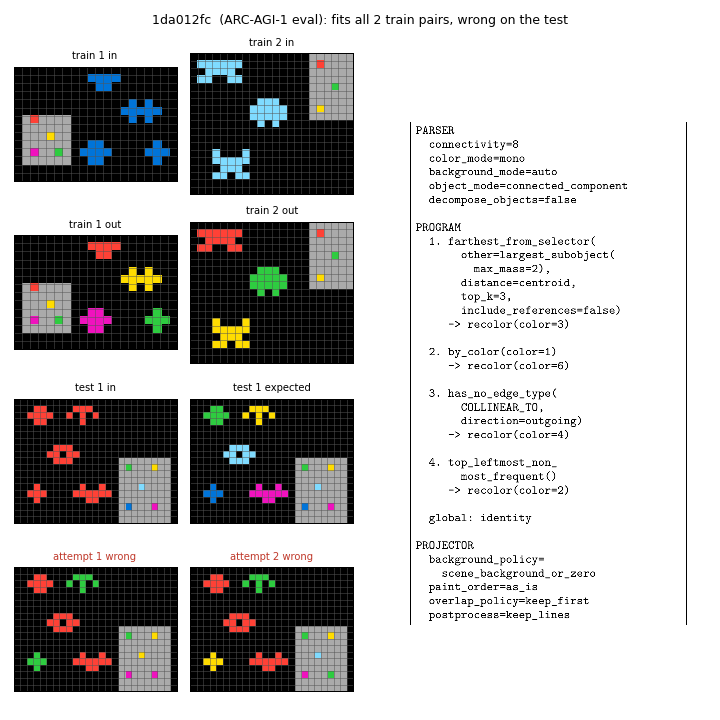}
    \par\smallskip
    \textbf{(c)} Representation limitation
\end{minipage}
\caption{Qualitative examples from ARC-AGI-1: (a) a successfully solved task; (b) a program fitting all demonstrations but failing on an unseen configuration; (c) a task requiring interpretation of a spatial color legend.}
\label{fig:qualitative-examples}
\end{figure*}

Task \texttt{4093f84a}, illustrated in Fig.~\ref{fig:qualitative-examples}(a), provides an example of a successfully evolved solution. The task requires moving colored objects toward a gray bar until they touch it, then recoloring them gray. The evolved parser configuration provides a suitable abstraction, representing the bar as the largest object and exposing the colored dots as manipulable subobjects. This allows the program to express the transformation through only two rules: the first slides all but the largest object toward the largest until blocked, while the second recolors all non-background objects gray. The resulting solution reproduces all three demonstrations and correctly predicts the test output at its first attempt, illustrating how the joint optimization of representations, transformations, and reconstruction can produce compact and effective solutions.

Task \texttt{2c608aff}, shown in Fig.~\ref{fig:qualitative-examples}(b), illustrates a limitation of generalization from a finite set of demonstrations. The evolved program selects isolated pixels and extends rays toward other objects until blocked, correctly reproducing all four training outputs. However, the test input introduces additional pixels aligned with these rays, causing unintended connections. Since no training example contains this configuration, the inferred rule is consistent with all available evidence, raising questions about the ambiguity of the intended generalization in ARC tasks. Even if the prediction were identified as incorrect, determining the intended alternative would remain a separate challenge, motivating further investigation into mechanisms for reassessing candidate solutions.

A different limitation appears in task \texttt{1da012fc}, shown in Fig.~\ref{fig:qualitative-examples}(c). Solving this task requires interpreting the gray region as a color legend and establishing a correspondence between the arrangement of its markers and the objects in the grid. Although the evolved program fits both demonstrations through a sequence of object selections and recoloring operations, it does not capture this underlying relationship. The current parser and \gls{dsl} lack a direct mechanism for representing and applying such context-dependent correspondences, which could otherwise require a particularly complex and unlikely combination of rules. This example illustrates how limitations in the available symbolic representations and operations can restrict the solver independently of its search capabilities.

\FloatBarrier

\section{Conclusions and Future Work}
\label{sec:discussion}

We presented a purely symbolic evolutionary approach to ARC-AGI that jointly optimizes the parser, symbolic program, and projector, using an object-aware fitness function to guide the search. Despite relying entirely on evolutionary computation, without neural components or pretrained models, our system achieves encouraging results, positioning itself among the leading approaches in the area of symbolic program synthesis. While its performance remains below that of the strongest ARC-AGI solvers, these results demonstrate the potential of combining evolutionary search with adaptable symbolic representations.

Our experiments demonstrate the effectiveness of the proposed approach on a substantial portion of ARC-AGI-1, with ablation studies highlighting the contribution of fitness-guided evolutionary search. The results also show that relatively simple symbolic programs can successfully capture and generalize the intended transformations. ConceptARC proved valuable for identifying missing conceptual capabilities and guiding targeted extensions, ultimately achieving near-complete coverage and helping improve performance on ARC-AGI.

Future work will investigate the systematic optimization of the DSL and the parser and projector configuration spaces, including the minimization of redundant operations and the generalization of overly specialized ones. We hypothesize that such improvements could bring ARC-AGI-1 closer to saturation, while ARC-AGI-2 may require greater emphasis on compositionality and library learning, enabling the construction and reuse of increasingly complex abstractions. Another direction is to investigate a learned fitness estimator that, given an input and a candidate transformation, predicts the number of mutations separating it from a correct solution, using progressively mutated correct programs as training examples. These directions remain experimental, and their effectiveness will need to be established empirically. Finally, we intend to explore whether the same principles of symbolic representation and evolutionary search can be extended to interactive reasoning environments such as ARC-AGI-3.

\printbibliography

\vfill

\end{document}